\documentclass[letterpaper, 10 pt, conference]{ieeeconf}  

\usepackage{threeparttable}
\usepackage[T1]{fontenc}
\usepackage{amsmath,bm}
\usepackage{textcomp}
\usepackage{amsfonts}
\usepackage{graphicx}
\usepackage{subfigure}
\usepackage{booktabs}
\usepackage{multirow}
\usepackage{xcolor}
\usepackage{slashed}

\usepackage{amssymb}
\usepackage{mathtools}

\usepackage{amsthm}

\usepackage{hyperref}

\def\cD{{\mathcal{D}}}
\def\cA{{\mathcal{A}}}

\def\cL{{\mathcal{L}}}

\newcommand{\bai}[1]{{\color{black}{#1}}}
\definecolor{darkorange}{rgb}{1.0, 0.55, 0.0}
\newcommand{\xinyi}[1]{{\color{black}{#1}}}
\newcommand{\xinyinew}[1]{{\color{black}{#1}}}

\IEEEoverridecommandlockouts                              

\title{\large \bf
Preference Aligned Diffusion Planner for \\ Quadrupedal Locomotion Control
}

\author{Xinyi Yuan$^{*1}$, Zhiwei Shang$^{*2}$, Zifan Wang$^{2}$, Chenkai Wang$^{4}$, Zhao Shan$^{3}$, \\Meixin Zhu$^{\dag5}$, Chenjia Bai$^{\dag3}$, Weiwei Wan$^{1}$, Kensuke Harada$^{1}$, Xuelong Li$^{3}$
\thanks{$^{*}$Equal Contribution.}
\thanks{$^{\dag}$Corresponding Authors: Meixin Zhu (e-mail: meixin@ust.hk), Chenjia Bai (e-mail: baichenjia255@gmail.com)}
\thanks{$^{1}$ Graduate School of Engineering Science, Osaka University, Japan.}%
\thanks{$^{2}$ Robotics and Autonomous Systems Thrust, The Hong Kong University of Science and Technology (Guangzhou), China.}%
\thanks{$^{3}$ Institute of Artificial Intelligence (TeleAI), China Telecom, China}%
\thanks{$^{4}$ Department of Statistics and Data Science, Southern University of Science and Technology, China}%
\thanks{$^{5}$ School of Transportation, Southeast University, China}
}

\begin{document}

\maketitle
\thispagestyle{empty}
\pagestyle{empty}

\begin{abstract}
Diffusion models demonstrate superior performance in capturing complex distributions from large-scale datasets, providing a promising solution for quadrupedal locomotion control. 
\xinyi{However, the robustness of the diffusion planner is inherently dependent on the diversity of the pre-collected datasets. To mitigate this issue, we propose a two-stage learning framework to enhance the capability of the diffusion planner under \xinyinew{limited dataset (reward-agnostic)}.} Through the offline stage, the diffusion planner learns the joint distribution of state-action sequences from expert datasets without using reward labels. Subsequently, we perform the online interaction in the simulation environment based on the trained offline planner, which significantly \xinyi{diversified the original behavior and thus improves the robustness}. Specifically, we propose a novel \emph{weak} preference labeling method without the ground-truth reward or human preferences. The proposed method exhibits superior stability and velocity tracking accuracy in pacing, trotting, and bounding gait under \xinyi{different speeds} and can perform a zero-shot transfer to the real Unitree Go1 robots. The project website for this paper is at \url{https://shangjaven.github.io/preference-aligned-diffusion-legged/}.

\end{abstract}

\section{INTRODUCTION}
Learning-based approaches significantly enhance the agility and adaptability of quadrupedal robots to accomplish diverse \bai{locomotion} tasks~\cite{zhuang2023robot,hwangbo2019learning}. While online learning demonstrates robustness in complex dynamic environments, extensive trial-and-error interactions in simulation are required to learn an effective policy. \bai{Thus, learning an online policy can be sample inefficient and requires a meticulously designed reward function.}
In contrast, offline learning can leverage the advantages of pre-collected offline datasets \bai{via model-based controller \cite{villarreal2020mpc}, animal imitation \cite{AMP}, or Reinforcement Learning (RL) policy \cite{RMA},} significantly improving data efficiency and reducing the cost of online interactions \cite{nie2022data}. \bai{In offline policy learning, diffusion models \cite{sohl2015deep} have shown superior performance in capturing complex action distributions from offline trajectories \cite{zhu2023diffusion}, \xinyi{exhibit promising potential} to solve quadruped locomotion tasks with high-dimensional action space and complex action distribution in various terrains.} \bai{As an example, DiffuseLoco \cite{huang2024diffuseloco} has recently proposed to train a diffusion planner for quadrupedal locomotion from \xinyi{large-scale} offline trajectories.}

\begin{figure}[!t]
    \centering
    \includegraphics[width=\linewidth]{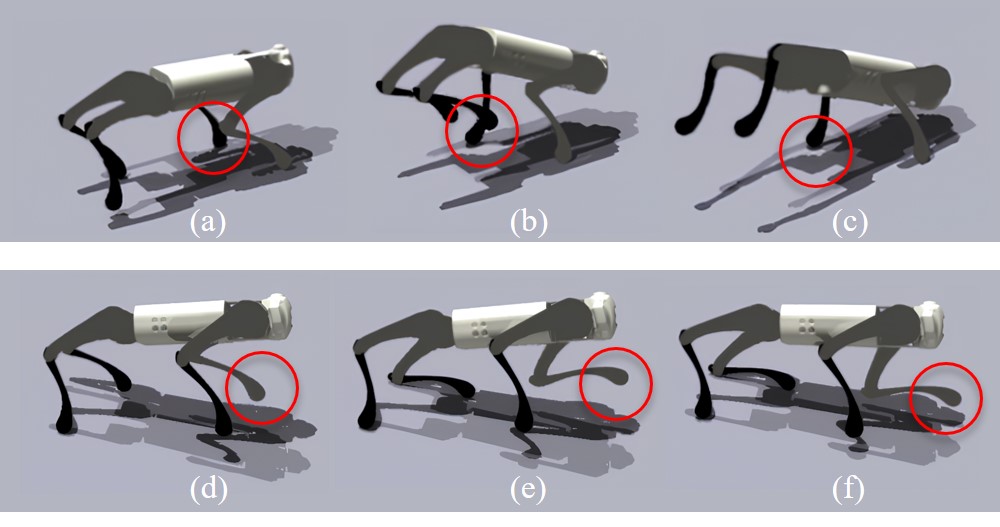}
    \caption{Failure Cases of the Offline Diffusion Planner under a Single-Source Limited Dataset: (a-c) Bounding Gait Example: The lift delay of the highlighted leg (red circle) deteriorates in subsequent motion sequences, ultimately causing the legged robot to fall. (d-f) Pacing Gait Example: The highlighted leg (red circle) demonstrates a higher lifting compared to the average step height, leading to severe lateral tilting.}
    \label{fig:fail_case}
\end{figure}

However, learning robust locomotion planners in a purely offline manner requires collecting a large-scale dataset with wide state coverage \cite{chang2021mitigating}, especially covering situations that the robot may encounter in the real world. \xinyinew{This exposes significant challenges in quadrupedal locomotion scenario, as it's difficult to comprehensively cover gait variations and diverse terrain dynamics.} Therefore, it is challenging to learn a robust locomotion policy using the \xinyi{limited} offline dataset with \xinyi{inadequate} state coverage, as the learned policy may be sensitive to out-of-distribution (OOD) states in the real world. 
\xinyinew{DiffuseLoco \cite{huang2024diffuseloco} addressed the challenge by leveraging a large-scale multi-skill dataset incorporating both quadrupedal and bipedal locomotion strategies from various source policies. The comprehensive state coverage leads to robust offline planner, enabling successful sim-to-real transfer.} \xinyinew{Rather than extensively increasing the dataset distribution, we aim to address the aforementioned issue by combining offline diffusion modeling with online interactions, where the collection of online trajectories augments the coverage of the state-action space distribution to align with the real world.  Notably, unlike previous online refinement algorithm that requires real-time action or reward labels, our proposed framework eliminates the need for external expert labels or reward functions.}

\begin{figure*}[!t]
\centering
\subfigure[Trotting Gait Simulation]{\includegraphics[width=0.4\linewidth]{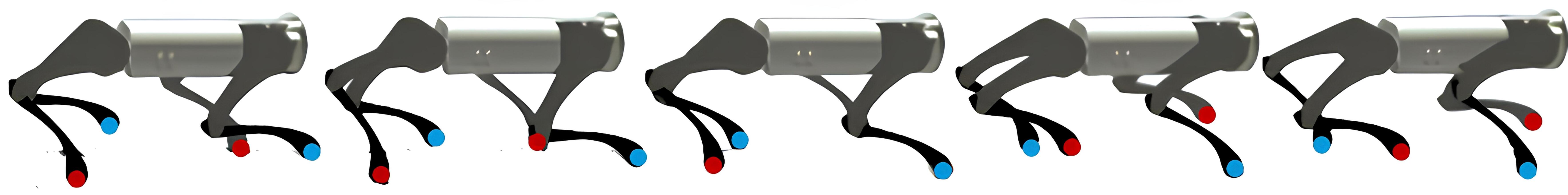}}
\hspace{0.05\linewidth}
\subfigure[Trotting Gait Real Robot]{\includegraphics[width=0.45\linewidth]{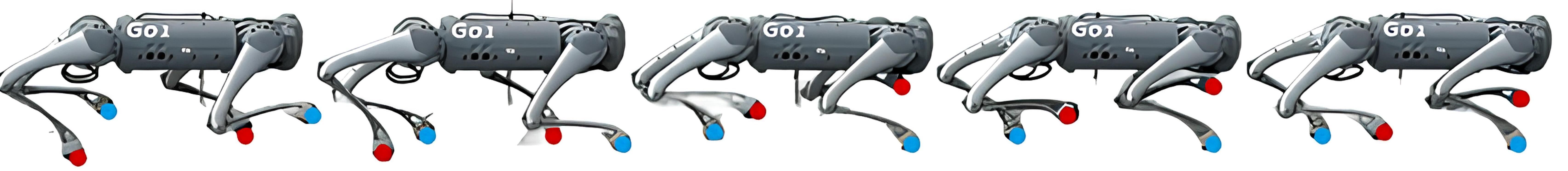}}\\ 
\subfigure[Pacing Gait Simulation]{\includegraphics[width=0.4\linewidth]{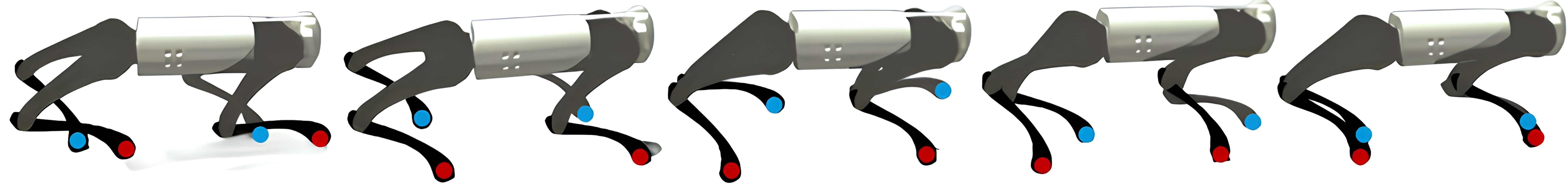}}
\hspace{0.05\linewidth}
\subfigure[Pacing Gait Real Robot]{\includegraphics[width=0.45\linewidth]{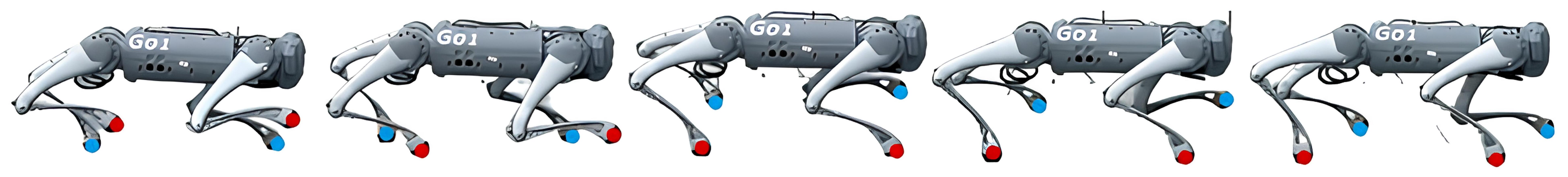}}\\
\subfigure[Bounding Gait Simulation]{\includegraphics[width=0.4\linewidth]{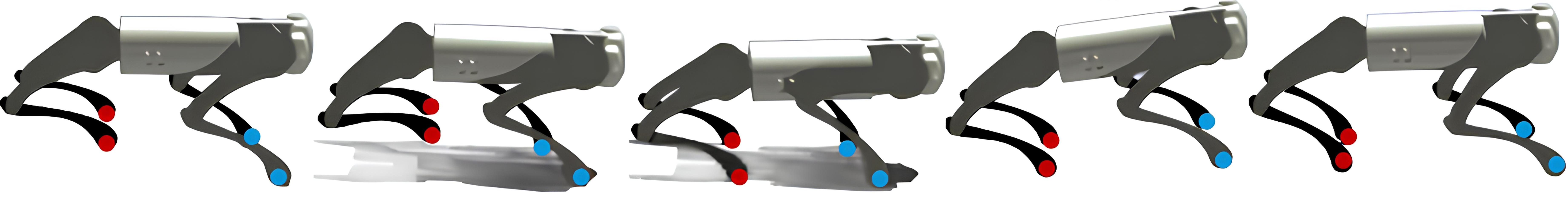}}
\hspace{0.05\linewidth}
\subfigure[Bounding Gait Real Robot]{\includegraphics[width=0.45\linewidth]{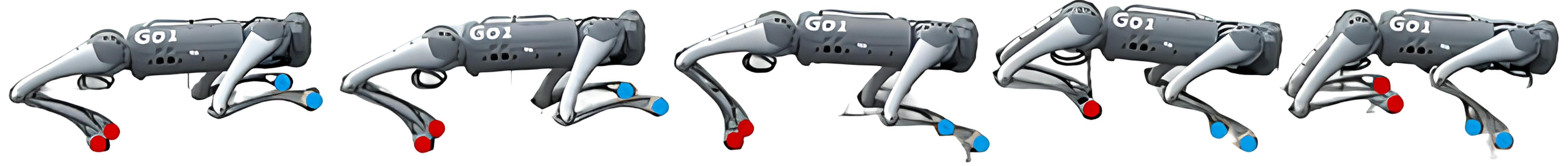}}

\caption{Video Frames in Simulation and Real World Experiments of the Proposed Architecture: (a-b) Trotting gait simulation and real-world test, (c-d) Pacing gait simulation and real-world test, (e-f) Bounding gait simulation and real-world test}
\label{fig:case}
\end{figure*}

In this paper, we propose a two-stage learning framework combining offline and online learning for legged locomotion. In the offline learning stage, the diffusion planner \bai{learns the joint distribution of state-action sequence} from an \xinyi{limited} offline dataset collected by \xinyi{a single pre-trained source policy} and does not use reward labels in training. Then, we perform online interactions with the environment based on
the diffusion planner, \xinyi{enhancing the diversity of the state distribution} and improving robustness via preference alignment. Specifically, we propose a novel online fine-tuning algorithm for the diffusion planner based on preferences, which resembles Direct Preference Optimization (DPO) \cite{an2023direct} \bai{for diffusion models in text-to-image generation
\cite{wallace2024diffusion}. Importantly, the preference score in our method is measured by distances between the states and the nearest neighbors of expert trajectories, which signifies the optimality of a trajectory and is used as the metric to construct preference pairs. As a result, the preference data can be easily constructed from such a \emph{weak} preference label without the ground-truth reward function or human preferences.} 

The contributions are summarized as follows: (\romannumeral1) We propose a novel two-stage learning framework that combines online interaction and offline diffusion learning \xinyi{to enhance the diversity and robustness of the diffusion planner under the limited expert dataset}; (\romannumeral2) We give an efficient preference alignment algorithm for offline diffusion planner via DPO and \emph{weak} preference labels; (\romannumeral3) The resulting diffusion planner exhibits superior performance on stability and velocity tracking accuracy in simulation and can be zero-shot transferred to the real-world Unitree robot.

\section{Related Work}
\subsection{Learning-Based Approaches for Legged Locomotion}
For legged locomotion, learning-based methods automatically capture dynamic behaviors from interacting experiences, largely reducing the need for manual expertise \bai{in classical control} \cite{lee2024learning}. Online RL has been widely applied in learning complex locomotion skills in simulation, and adaptive techniques like domain randomization are employed to facilitate the transfer from the simulation to real robot \cite{margolis2023walk,peng2016terrain,schneider2024learning,miki2022learning,rudin2022learning}. However, manually designing reward functions and tuning weights can be particularly challenging when dealing with complex tasks. Other methods \cite{peng2020learning} adopt imitation learning to extract agile locomotion strategies from real-world animal reference motions, \bai{while it requires the motion-capture dataset that is more expensive \cite{han2024lifelike}.} Recent studies utilized offline learning on locomotion control with the limited scope confined to gym simulation environments \cite{levine2020offline,torabi2018behavioral}. \xinyi{While DiffuseLoco \cite{huang2024diffuseloco} demonstrated the strong performance of diffusion locomotion planner, it requires large-scale datasets among various source policies (i.e., three different RL-based methods with 14 million data in total).}

\subsection{Diffusion Models in Robotics}

Diffusion models have demonstrated superior generative capabilities in various robotics tasks such as robotic navigation \cite{sridhar2024nomad}, manipulation \cite{mishra2024reorientdiff,scheikl2024movement}, and decision-making \cite{ajay2022conditional}. For example, Diffuser \cite{janner2022planning} is a planner that analogizes the planning to the denoising process in diffusion models, demonstrating impressive adaptability in complex long-horizon manipulation tasks. Recent studies \cite{pearce2023imitating, chi2023diffusion} represent robot policies also as the diffusion process, where the policy generates joint actions based on multi-modal conditional inputs such as observations or visual information. However, most existing research has been limited to high-level tasks with low-dimensional action spaces, leaving room for exploration in more complex, high-dimensional scenarios such as legged locomotion.

\section{two-stage learning framework}
\subsection{System Design}
\begin{figure*}[!t]
    \centering
    \includegraphics[width=1.0\linewidth]{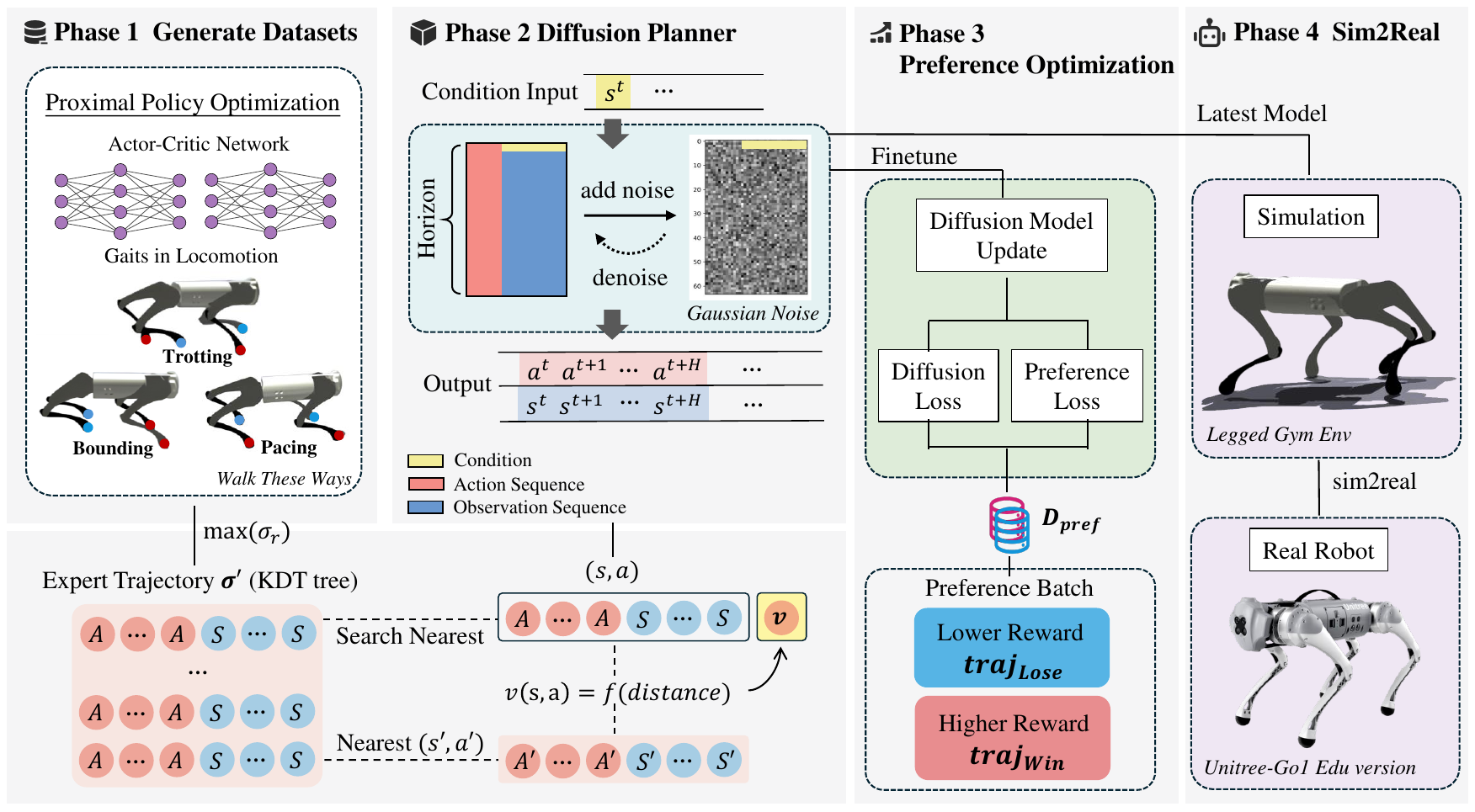}
    \caption{Proposed Architecture Framework Overview: \textbf{(1) Generate Datasets}: the offline datasets among pacing, trotting, and bounding gait are collected through the expert PPO policy in the $\textit{walk-these-ways}$ task. \textbf{(2) Behavior Cloning}: given a condition input $s^t$, the diffusion policy can produce a sequence of states and actions. \textbf{(3) Preference Alignment}: Conduct the preference alignment on the offline diffusion planner based on proposed \textit{weak} preference labels. \textbf{(4) Sim2Real}: The refined policy is deployed on the Unitree Go1 robot.}
    \label{fig:system}
\end{figure*}

The architecture of the proposed framework is illustrated in Fig. \ref{fig:system}. The entire system consists of four stages. Firstly, preference-free offline datasets (i.e., $\cD$) containing various gait patterns are collected in the \textit{Walk-These-Ways} environment. Secondly, behavior cloning is performed by extracting gait-specific diffusion policies from these datasets. Following this, the diffusion policy is fine-tuned using the direct preference optimization method on the constructed preference dataset (i.e., $\cD_{\rm pref}$). Finally, the latest model will be deployed on the Unitree Go1 robot.

\subsection{Offline Dataset Generation}
This work utilizes \textit{Walk These Ways} framework \cite{margolis2023walk} as the source locomotion policy and collects offline datasets for the following gaits: pacing, trotting, and bounding. For each gait, we roll out 2048 episodes with 250 time steps. The data format as $(\textbf{s}_t, \textbf{a}_t,\textbf{s}_{t+1}, \textbf{r}_t, \textbf{d}_t)$, where the $\textbf{d}_t$ turns True when the episode ends. \bai{The behavior policy is denoted as} $\pi(\textbf{a}_t|\textbf{c}_t, \textbf{b}_t)$, 
where $\textbf{c}_t$ includes commands, and $\textbf{b}_t$ indicates the behavior parameters. The reward functions of the \xinyi{source} locomotion controller include task reward (tracking the command speed), augmented auxiliary (task behavior related), and fixed auxiliary (promoting robot stability).

The Unitree Go1 robot possesses 12 degrees of freedom, with each leg comprising 3 degrees of freedom, corresponding to the hip, thigh, and calf joints. The observation space $\textbf{O}_t$ of offline diffusion planner consists of the state vector $\textbf{s}_t = \{\textbf{v}^{cmd}_t, \textbf{q}_t, \dot{\textbf{q}_t}, \textbf{t}_t\} \in \mathbb{R}^{31}$. Specifically, $\textbf{v}^{cmd}_t \in \mathbb{R}^{3}$ incorporates the velocity command along x, y and z axis, $\textbf{q}_t \in \mathbb{R}^{12}$ represent the joint positions, $\dot{\textbf{q}_t} \in \mathbb{R}^{12}$ represent the joint velocities, $\textbf{t}_t \in \mathbb{R}^{4}$ is the gait and timing related parameter. Each feature of the observations is normalized by a Gaussian distribution before input into the diffusion models. 

The action space $\textbf{A}_t \in \mathbb{R}^{12}$ represents each joint's position targets. We follow the actuator network \cite{hwangbo2019learning} for actions-to-torques mapping. The legged robot will then execute the resultant torque.

\subsection{Offline Diffusion Planner}
\bai{We adopt a \xinyi{conditional} diffusion model to train an offline diffusion planner for locomotion.} The conditional input contains the current observation state $\textbf{s}_0$ (ensuring the generated trajectory starts from the current state), and the output is the action sequence $\textbf{a}_{0:h}$ of length horizon. 

\xinyi{The proposed framework uses \textit{Denoising Diffusion Probabilistic Models} (DDPM) \cite{ho2020denoising} with U-net and Transformer backbone. DDPM starts by adding Gaussian noise to the original data stepwise and uses the neural network to learn the inverse denoising process. During the forward process, Gaussian noise will be added gradually to the original data:
\begin{equation}
    q(\mathbf{x}_t|\mathbf{x}_{t-1}) = \mathcal{N}(\mathbf{x}_t; \sqrt{1-\beta_t}\mathbf{x}_{t-1}, \beta_t\textbf{I}),
\end{equation}
where $\beta_t$ is the variance scheduler, $\mathbf{x}_t$ and $\mathbf{x}_{t-1}$ are samples from two adjacent diffusion steps.


During the reverse chain, the model starts from the pure Gaussian noise $\mathbf{x}_T$ and gradually extracts the noise to derive $\mathbf{x}_{T-1}$ ... $\mathbf{x}_0$. DDPM uses $p_\theta(\mathbf{x}_{t-1}|\mathbf{x}_{t})$ to approximate the conditional denoising distribution:
\begin{equation}
\begin{aligned}
p_\theta(\mathbf{x}_{t-1} \vert \mathbf{x}_t) = \mathcal{N}(\mathbf{x}_{t-1}; \boldsymbol{\mu}_\theta(\mathbf{x}_t, t), \boldsymbol{\Sigma}_\theta(\mathbf{x}_t, t)).
\end{aligned}
\end{equation}

For training, we use the simplified objective without the weighting term \cite{ho2020denoising} as
\begin{equation}
    \begin{aligned}
        \mathcal{L}_t^\text{simple}(\theta)
        = \mathbb{E}_{\textbf{x}_0,\boldsymbol{\epsilon},t } \Big[\|\boldsymbol\epsilon_t - \boldsymbol{\epsilon}_\theta(\mathbf{x}_t, t)\|^2 \Big].
    \end{aligned}
\end{equation}}
During the training process, the data loader will randomly sample the normalized trajectory segment from the offline dataset with batch size $N$, and then Gaussian noise will be added iteratively into the trajectories. Subsequently, the diffusion model will denoise the noisy trajectory to reconstruct the original data. The diffusion loss measures the Mean-Squared Error (MSE) between the true noise and predicted noise:
\begin{equation}
\mathcal{L} = MSE(\boldsymbol{\epsilon}_k, \boldsymbol{\epsilon}_\theta(\textbf{a}_k^{t:t+h}, \textbf{s}^t_0, k))
\end{equation}
Notably, $\epsilon_\theta$ is the noise predictor, the superscript represents the time step in the trajectory (start from time $t$ with the horizon $h$), while the subscript denotes the diffusion denoising iterations ($k$). Besides, a fixed mask will be applied on the loss to ignore the deviations in the observation condition ($s_0$). 

During the inference, the Isaac Gym environment will be reset initially, and the current observation serves as the \xinyi{conditional} input for trajectory generation. Then, the trained diffusion model will denoise the pure Gaussian noise to derive the action sequence. Specifically, Classifier-Free Guidance (CFG) \cite{ho2022classifier} is employed to blend the conditioned and unconditioned predictions with the weight $w$:
\begin{equation}
    \bar{\boldsymbol{\epsilon}}_\theta(\mathbf{x}_t, t, \textbf{y}) = (1+w)\boldsymbol{\epsilon}_\theta(\mathbf{x}_t, t, \textbf{y}) - w \boldsymbol{\epsilon}_\theta(\mathbf{x}_t, t)
\end{equation}
where $\bar{\boldsymbol{\epsilon}}_\theta$ is the predictor under classifier-free guidance, $y$ represent the conditional information.

After generating the output action sequence of length horizon ($h$), the robot will execute every action stepwise and record the state transitions. The inference and interaction will continue until the episode length reaches the pre-defined maximum length. To ensure real-time performance, we decrease the diffusion sampling step in inference to 10 steps without sacrificing the locomotion performance. Besides, since the whole generated action sequence will be executed, the required inference frequency within each episode will be reduced compared with only implementing the single action.

\section{\xinyi{Preference Aligned Diffusion Planner}}

\xinyi{To enhance the state distribution diversity and policy robustness, we propose the preference alignment method to fine-tune the offline diffusion planner.} Firstly, we roll out preference-free datasets $\cD$ from the pre-trained offline diffusion planner. When constructing the preference dataset $\cD_{\rm pref}$, we randomly sample two segments from $\cD$ without replacement and then assign preference labels based on the following method. 

Considering that \bai{the ground-truth} rewards are sometimes challenging to obtain, we propose a \emph{weak} preference labeling method that allows for constructing preference labels without the reward labels, requiring only a few expert trajectories. The optimal expert trajectory is selected from the given expert trajectories based on their cumulative reward. For each state-action pair $(\textbf{s}, \textbf{a})$ in the preference-free dataset $\cD$, we search for the closest state-action pair $(\textbf{s}', \textbf{a}')$ in the \emph{optimal} expert trajectory and calculate their Euclidean distance. The value ($v$) of specific $(\textbf{s}, \textbf{a})$ is calculated as the equation:
\begin{equation}
    v_t = \exp\left(-\frac{\beta \times d_t}{|\cA|}\right)
\end{equation}
where $\beta$ is a hyper-parameter with a value of $0.5$, $d_t$ represents the calculated Euclidean distance, $|\cA|$ is the dimension of the action space. Notably, a similar technique was applied in \cite{lyu2024seabo}. However, we have modified this approach by querying only based on the current state and action.

Finally, given two trajectories $\boldsymbol{\sigma}_1$ and $\boldsymbol{\sigma}_2$, we compute the corresponding values for each state-action pair within them. The segment with a higher cumulative \emph{value} will be determined as the winning segment ($\boldsymbol{\sigma}_{\rm winning}$), as inferred by the following equation: 
\begin{equation}
\boldsymbol{\sigma}_{\text{winning}}^1 = \arg\max_{\boldsymbol{\sigma} \in \{\boldsymbol{\sigma}_1, \boldsymbol{\sigma}_2\}} \left( \sum_{t=0}^{h} v_t^{(1)}, \sum_{t=0}^{h} v_t^{(2)} \right)
\end{equation}

To demonstrate the effectiveness of our proposed \emph{weak} preference, we provide the alternative reward-available Preference Label (\emph{strong label}) for experiment comparison. Suppose the reward label can be obtained in the environment, and the segment with higher \emph{ cumulative reward} will be preferred, indicated in the following equation:
\begin{equation}
\boldsymbol{\sigma}_{\text{winning}}^2 = \arg\max_{\boldsymbol{\sigma} \in \{\boldsymbol{\sigma}_1, \boldsymbol{\sigma}_2\}} \left( \sum_{t=0}^{h} r_t^{(1)}, \sum_{t=0}^{h} r_t^{(2)} \right)
\end{equation}




According to the Bradley-Terry model, the preference model can be described as (here $\boldsymbol{\sigma}_{\text{winning}}$ is abbreviated as $\boldsymbol{\sigma}^+$, the $\boldsymbol{\sigma}_{\text{losing}}$ is abbreviated as $\boldsymbol{\sigma}^-$):
\begin{align}
    P(\boldsymbol{\sigma}^+>\boldsymbol{\sigma}^-) = \rm Sigmoid(r(\boldsymbol{\sigma}^+) - r(\boldsymbol{\sigma}^-))
\end{align}

The proposed loss function during the preference alignment incorporates two components: the preference loss ($\cL_{\rm DPO}$) derived from DPO and the regularization term. The preference loss amplifies the difference between the ``winning segment'' and the ``losing segment,'' thus improving the diffusion policy's performance to align with the preference labels and generate more preferred samples. More importantly, the regularization term helps to avoid significant deviations from the original policy. The loss function for the preference alignment stage is as follows:
\begin{align}
    \cL(\boldsymbol{\epsilon}_\theta, \cD_{\rm pref}) = \cL_{\rm DPO}(\boldsymbol{\epsilon}_\theta, \cD_{\rm pref}) - \mu \mathbb{E}_{\boldsymbol{\sigma} \in \cD} \log \boldsymbol{\epsilon}_\theta(\boldsymbol{\sigma})
\end{align}
The practical approximation and detailed proof can be found in \cite{shan2024forward}.

\section{Experimental Results}
\xinyi{We conduct the velocity tracking task among three gaits with different speeds to quantify the performance of our proposed framework. The diffusion pipeline is implemented under CleanDiffuser \cite{dong2024cleandiffuser}, hyperparameters are listed in Table \ref{tab:hyper}. For evaluation metrics, we consider the stability and average $x$-axis speed. We assess an episode as stable when the quadruped robot does not fall during all $250$ steps. For every experimental parameter setting, we repeatedly collect $1024$ episodes for three random seeds.}

\begin{table}[!h]
\centering
\caption{Hyper-parameters in Training, Inference, and Preference Alignment Stage}
\label{tab:hyper}
\begin{tabular}{@{}llc@{}}
\toprule
\textbf{Stage} & \textbf{Hyper-parameter} & \textbf{Value} \\ \midrule
\multirow{5}{*}{\begin{tabular}[c]{@{}l@{}}Offline Diffusion Planner\end{tabular}} & Batch size & 64 \\
 & Horizon & 64 \\
 & Solver & DDPM \\
 & Diffusion steps & 20 \\
 & Action loss weight & 10.0 \\ \midrule
\multirow{2}{*}{Inference} & w\_cg & 0.0001 \\
 & Sampling steps & 10 \\ \midrule
\multirow{3}{*}{Preference Alignment}
 & Regularization weight & 1.0 \\
 & Bias & 0 \\
 & Temperature & 500 \\ \bottomrule
\end{tabular}
\end{table}

\begin{table*}[ht]
\centering
\caption{Average Speed and Stability Comparison among Methods on Quadruped Locomotion Velocity Tracking Tasks}
\label{result}
\begin{threeparttable}
\resizebox{\linewidth}{!}{
\begin{tabular}{lccccccccc}
\toprule
 & \multicolumn{1}{c}{CQL} 
 & \multicolumn{2}{c}{\begin{tabular}[c]{@{}c@{}}Offline Planner\\ (DDPM-UNet)\end{tabular}} 
 & \multicolumn{2}{c}{\begin{tabular}[c]{@{}c@{}}Offline Planner\\ (DDPM-Transformer)\end{tabular}}
 & \multicolumn{2}{c}{\begin{tabular}[c]{@{}c@{}}\emph{Weak-Preference} \\Aligned Planner\end{tabular}}
 & \multicolumn{2}{c}{\begin{tabular}[c]{@{}c@{}}Reward-Preference \\Aligned Planner\end{tabular}} \\
\cmidrule(lr){2-2} \cmidrule(lr){3-4} \cmidrule(lr){5-6} \cmidrule(lr){7-8} \cmidrule(lr){9-10}
& $-$ & Speed (m/s) & Stability (\%) & Speed (m/s) & Stability (\%) & Speed (m/s) & Stability (\%) & Speed (m/s) & Stability (\%) \\
\midrule
Pacing (0.5 m/s) & fail & 0.60 & 40.9 & 0.43 & 72.0 & 0.46 & 81.6 & \textbf{0.49} & \textbf{87.3} \\
Trotting (0.5 m/s) & fail & 0.59 & 39.5 & 0.36 & 58.7 & 0.42 & 78.5 & \textbf{0.50} & \textbf{85.2} \\
Bounding (0.5 m/s) & fail & 0.59 & 40.4 & 0.21 & 60.1 & 0.44 & \textbf{70.8} & \textbf{0.48} & \textbf{72.4} \\
\midrule
Pacing (1.0 m/s) & fail & 0.49 & 77.4 & 0.62 & 81.4 & 0.64 & \textbf{91.6} & \textbf{0.73} & \textbf{91.8} \\
Trotting (1.0 m/s) & fail & 0.48 & 58.9 & 0.61 & 66.6 & 0.63 & 84.4 & \textbf{0.67} & \textbf{89.5} \\
Bounding (1.0 m/s) & fail & 0.42 & 32.4 & 0.78 & 46.6 & 0.72 & 84.3 & \textbf{0.80} & \textbf{89.2} \\
\bottomrule
\end{tabular}}
\end{threeparttable}
\end{table*}

We chose three strong baselines for the experiments. The first one is \textbf{Conservative Q-learning (CQL)}. CQL mitigates the Q-value overestimation problem by introducing the regularization term in conservative Q-value estimation. CQL is effective in high-dimensional state space and complex environments. The second and third baselines are offline diffusion planners with U-Net backbone and transformer backbone, respectively. They are noted as \textbf{DDPM-Unet} and \textbf{DDPM-Transformer}.


\subsection{Performance of Preference-aligned Diffusion Planner}
Table \ref{result} presents a comparison of the proposed method with baselines across different gaits (pacing, trotting, bounding) and speeds (0.5 m/s and 1.0 m/s) in terms of stability and average $x$-axis velocity.

CQL consistently failed all the locomotion tasks. We observed in the simulation environment that the quadruped robot either remained stationary on the ground or exhibited subtle irregular jitters. This indicates that CQL struggles with continuous control in complex locomotion tasks. 

\textbf{Stability Performance}: The proposed preference-aligned diffusion planner outperformed all baselines in stability across all locomotion tasks. Specifically, for the \xinyi{1.0 m/s} bounding task, the proposed \emph{weak} preference-aligned planner achieves 84.3\% stability, much higher than 32.4\% of the DDPM-Unet and 46.6\% of the DDPM-Tranformer. Additionally, in the \xinyi{0.5m/s} trotting task, the stability increased by 19.8\% relative to the offline DDPM-Transformer model. These results indicate that \xinyi{our preference-aligned planner exhibits superior stability performance across different speeds among gaits}.

\xinyi{Notably, the improvement in stability from the DDPM-Transformer to the preference-aligned planner is attributed to the online alignment stage, which mitigates the inadequate diversity caused by a limited single-source offline dataset and aligns the planner with real-world data distribution. Experiments indicate that the offline diffusion planner in Fig. \ref{fig:system} Phase $2$ can generate reasonable action sequences, however, it is extremely sensitive to the OOD states encountered in the simulation environment, examples shown in Fig. \ref{fig:fail_case}. This is due to the high-quality expert locomotion trajectories did not enable the diffusion planner to effectively address non-fatal disturbances, the deviated action prediction will gradually be amplified to significant cumulative errors (i.e., falls).}

\textbf{Velocity Tracking Performance}: We evaluate the velocity tracking performance by the difference between the average measured $x$-axis velocity and the target velocity, and a more minor deviation indicates more accurate velocity control. For example, in the \xinyi{1.0 m/s} bounding task, our proposed planner achieves an average velocity of 0.72 m/s, closely approaching the target speed of 1.0 m/s, significantly outperforming DDPM-Unet (0.42 m/s). In the \xinyi{0.5 m/s} task, the proposed method demonstrates more precise velocity control with minor deviation from the target speed. For example, in the \xinyi{0.5 m/s} trotting task, our method achieved an average velocity of 0.42 m/s, closely matching the target and surpassing DDPM-Unet (0.59 m/s) and DDPM-Transformer (0.36 m/s).

\begin{table*}[!t]
\centering
\caption{ablation studies on preference alignment parameters}
\label{tab:prefer}
\begin{tabular}{@{}llccclccc@{}}
\toprule
 &  & \multicolumn{3}{c}{Speed = 1.0 m/s} & \textbf{} & \multicolumn{3}{c}{Speed = 0.5 m/s} \\ \cmidrule(lr){3-5} \cmidrule(l){7-9} 
 &  & Pacing & Trotting & Bounding &  & Pacing & Trotting & Bounding \\ \midrule
\multirow{3}{*}{Preference Number} & 1024 episodes (*0.5) & \textbf{89.6} & 78.8 & 75.3 &  & 69.5 & 73.0 & 63.5 \\
 & 2048 episodes (*1.0) & \textbf{90.1} & 77.7 & 75.9 &  & 76.4 & 79.1 & 66.4 \\
 & 3072 episodes (*1.5) & \textbf{91.8} & \textbf{89.5} & \textbf{89.2} &  & \textbf{87.3} & \textbf{85.2} & \textbf{72.4} \\ \midrule
\multirow{2}{*}{Preference Quality} & \emph{Weak} Preference Label & \textbf{91.6} & 84.4 & 84.3 &  & 81.6 & 78.5 & \textbf{70.8} \\ 
 & Strong Preference Label & \textbf{91.8} & \textbf{89.5} & \textbf{89.2} &  & \textbf{87.3} & \textbf{85.2} & \textbf{72.4} \\ \midrule
\multirow{2}{*}{Regularization} & Without Regularization & $-$ & $-$ & $-$ &  & $-$ & $-$ & $-$ \\
 & With Regularization & \textbf{91.8} & \textbf{89.5} & \textbf{89.2} &  & \textbf{87.3} & \textbf{85.2} & \textbf{72.4} \\ \bottomrule
\end{tabular}
\end{table*}

\begin{figure}[!h]
    \centering
    \includegraphics[width=0.9\linewidth]{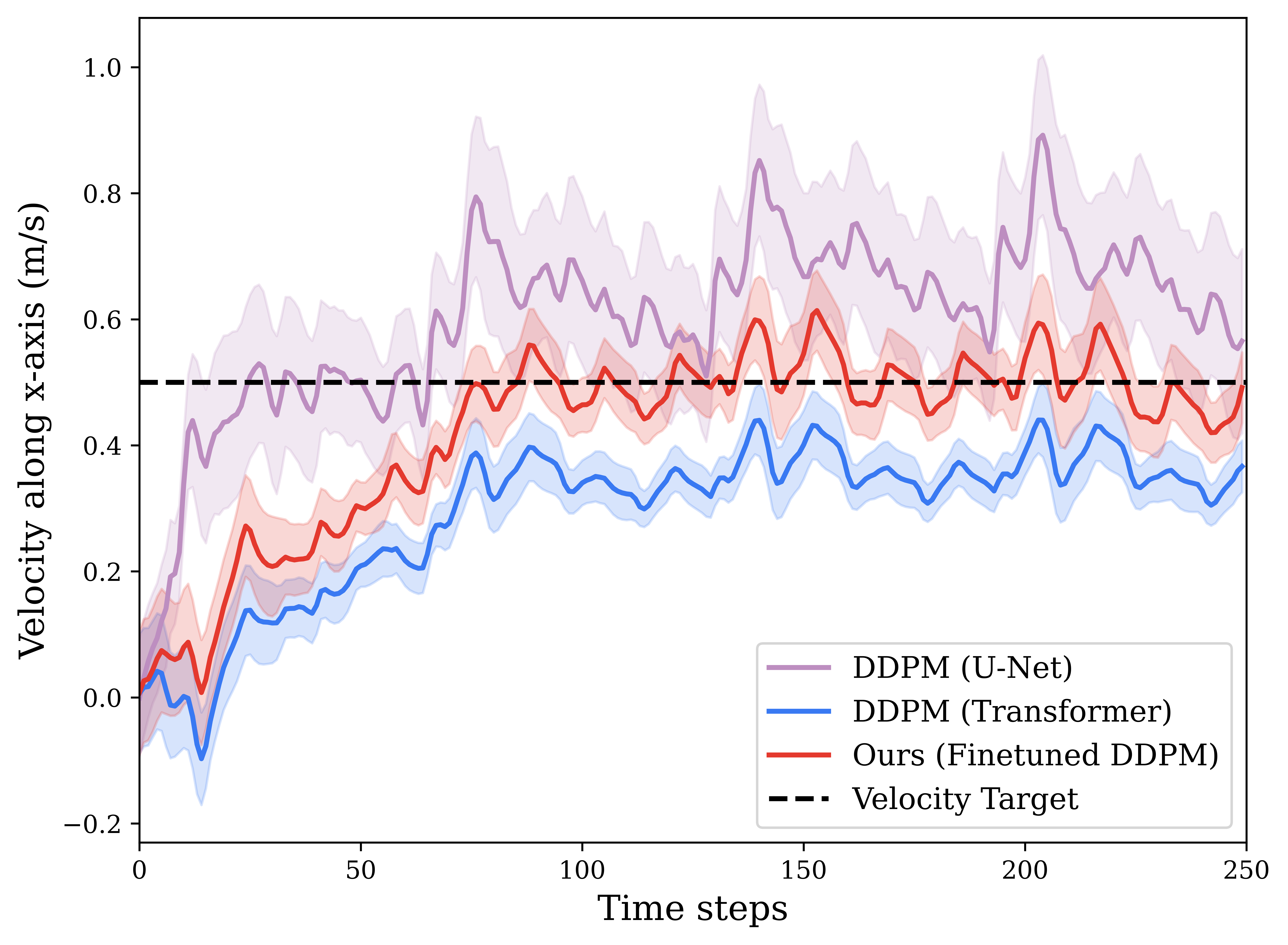}
    \caption{Velocity tracking result in \xinyi{0.5 m/s} bounding gait between models}
    \label{fig:x_vel}
\end{figure}

To further present how our proposed method improves the velocity tracking performance, we conduct a case study based on the most challenging \xinyi{0.5 m/s} bounding task with the lowest stability in Fig. \ref{fig:x_vel}.

Fig. \ref{fig:x_vel} depicts the velocity tracking process within an episode; the real-time velocities are smoothed with a moving average filter. The DDPM-Transformer exhibits a smaller standard deviation while it still deviates from the target velocity. Additionally, DDPM-Unet shows high fluctuation in measured velocity and fails to track the desired velocity. In contrast, the proposed preference-aligned diffusion planner (red line) reaches and maintains close to the target velocity.

In conclusion, the experimental result demonstrates that the proposed method outperforms existing baselines on stability and velocity tracking performance. The quantitative analysis supports the effectiveness and robustness of our proposed two-stage learning framework \xinyi{among different locomotion tasks}.

\subsection{Real-world Experiments}
\begin{figure}[b]
    \centering
    \includegraphics[width=1\linewidth]{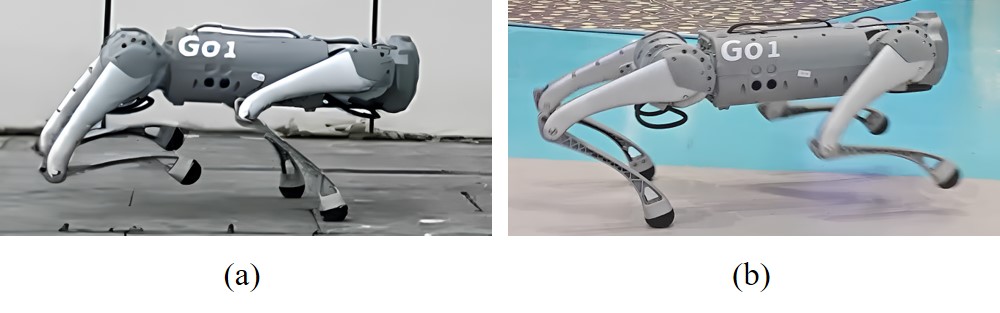}
    \caption{Bounding Gait Testing on Different Surfaces: (a) Cement Surface which is Rough and High Friction, (b) PVC Floor which is Smooth and Low Friction)}
    \label{fig:real_exp}
\end{figure}

To evaluate the robustness and adaptability of the locomotion policy derived by the proposed two-stage learning framework, real-world experiments were conducted on two surfaces with different frictional properties: a rough, high-friction cement surface and a smooth, low-friction PVC floor. Fig. \ref{fig:real_exp} presents key frames during the takeoff and landing phases of the legged robot in the bounding gait, exhibiting foot-terrain interactions across different surfaces. Experiment results indicate that the proposed framework can lead to a robust diffusion-based locomotion policy capable of sim-to-real transfer, even with a limited dataset.

\subsection{Ablation Studies}
The ablation studies in Table \ref{tab:prefer} investigate the impact of preference dataset size, preference label quality, and regularization methods on stability performance.

\textbf{Preference Dataset Size}: Increasing the preference dataset size from 1024 episodes to 3072 episodes improves the performance across all gaits and speeds. For example, in the \xinyi{0.5 m/s} pacing task, the stability increased by 17.8\%. Generally, the sensitivity to the size of the preference dataset varies with the difficulty of different locomotion tasks. In the simplest \xinyi{1.0 m/s} pacing task, 1024 episodes of the preference dataset are sufficient to achieve relatively stable improvement. We observe in the simulation result that a smaller preference dataset may introduce the risk of insufficient coverage on state distribution, \xinyi{leads to deviating from the expected task performance}. However, the overall results indicate our proposed preference alignment framework demonstrates stable improvement in the offline diffusion planner under limited preference data.

\textbf{Preference Label Quality}: We compare the performance between the reward-based preference label (strong preference) and our proposed reward-unavailable preference label (\emph{weak} preference). The results in Table \ref{tab:prefer} indicate that weak preference labels can achieve \xinyi{satisfactory} performance across most locomotion tasks with slight differences compared with strong preference labels. For example, the performance of the weak preference label closely approaches the strong preference label in the \xinyi{1.0 m/s} pacing and \xinyi{0.5 m/s} bounding tasks. Notably, the preference alignment based on weak preference labels demonstrates significant improvement compared with the DDPM-Transformer on all locomotion tasks, \xinyi{indicating the effectiveness of the weak preference labeling method and proposed two-stage learning framework}.

\textbf{Regularization Methods}: The ablation results in Table \ref{tab:prefer} suggest the critical role of the proposed regularization term in the preference alignment stage. The \xinyi{compared baseline method} eliminates the regularization term and solely emphasizes the difference between the winning segment ($\sigma_{\rm winning}$) and the losing segment ($\sigma_{\rm losing}$) under the current policy and reference policy. The results demonstrate that without the regularization term, the preference alignment will consistently cause \emph{fail} in all locomotion tasks. In comparison, our proposed preference alignment method encourages the generation of winning segments while maintaining the overall likelihood of the winning and losing segments through the regularization term, thus effectively \xinyi{constrains the step size of policy updates to ensure stable and conservative improvements within the online alignment stage.}

\section{CONCLUSIONS}
\xinyi{This paper presents a two-stage learning framework that integrates offline diffusion learning with online preference alignment to enhance the diversity and robustness of the diffusion planner. We focus on the challenging scenario with only a single-source limited expert dataset available and validate the effectiveness of the proposed architecture on the locomotion velocity tracking task.} We leverage the offline diffusion planner to approximate the complex state-action sequences and further utilize the proposed \emph{weak} preference label to conduct the preference alignment. Experiments indicate that our framework improved the stability and velocity tracking accuracy and can be deployed on Unitree Go1 robots. Future work can incorporate extra modalities, such as vision and language, into the learning framework. 

\section*{ACKNOWLEDGMENT}
This study is supported by the National Natural Science Foundation of China under Grant No.52302379 and Grant No.62306242. Acknowledgment to Computing Lab, Hong Kong University of Science and Technology (Guangzhou) for providing the computational resources utilized in this research.

\bibliographystyle{IEEEtran}
\bibliography{IEEEexample}

\end{document}